\documentclass[sigconf]{acmart}

\AtBeginDocument{%
  }

\usepackage[utf8]{inputenc}
\usepackage{microtype}
\usepackage{multirow}
\usepackage{subfigure}
\usepackage[utf8]{inputenc}
\usepackage{graphicx}
\usepackage{amsmath}
\usepackage{algorithm}
\usepackage{algorithmic}
\usepackage{CJKutf8}
\usepackage{balance}
\usepackage[skip=2.5pt]{caption}
\usepackage{array}
\usepackage{booktabs}
\usepackage{threeparttable}
\usepackage{soul}
\usepackage{setspace}

\setcopyright{acmlicensed}

\copyrightyear{2025} 
\acmYear{2025} 
\setcopyright{acmlicensed}\acmConference[SIGIR '25]{Proceedings of the 48th International ACM SIGIR Conference on Research and Development in Information Retrieval}{July 13--18, 2025}{Padua, Italy}
\acmBooktitle{Proceedings of the 48th International ACM SIGIR Conference on Research and Development in Information Retrieval (SIGIR '25), July 13--18, 2025, Padua, Italy}
\acmDOI{10.1145/3726302.3731952}
\acmISBN{979-8-4007-1592-1/2025/07}

\begin{document}

\title{
    Multi-objective Aligned Bidword Generation Model for E-commerce Search Advertising
}


\author{Zhenhui Liu}
\authornote{Both authors contributed equally to this research.}
\email{leon0425@stu.pku.edu.cn}
\author{Chunyuan Yuan}
\authornotemark[1]
\authornotemark[2]
\email{chunyuany93@outlook.com}
\affiliation{
  \institution{JD.COM}
  \city{Beijing}
  \country{China}
}

\author{Ming Pang}
\authornote{Authors are corresponding authors.}
\email{pangming8@jd.com}
\author{Zheng Fang}
\email{fangzheng21@jd.com}
\affiliation{
  \institution{JD.COM}
  \city{Beijing}
  \country{China}
}

\author{Li Yuan}
\authornotemark[2]
\email{yuanli-ece@pku.edu.cn}
\affiliation{
  \institution{Peking University}
  \city{Shenzhen}
  \country{China}
}

\author{Xue Jiang}
\email{jiangxue@jd.com}
\author{Changping Peng}
\email{pengchangping@jd.com}
\affiliation{
  \institution{JD.COM}
  \city{Beijing}
  \country{China}
}

\author{Zhangang Lin}
\email{linzhangang@jd.com}
\author{Zheng Luo}
\email{lawching@jd.com}
\affiliation{
  \institution{JD.COM}
  \city{Beijing}
  \country{China}
}

\author{Jingping Shao}
\email{shaojingping@jd.com}
\affiliation{
  \institution{JD.COM}
  \city{Beijing}
  \country{China}
}

\renewcommand{\shortauthors}{Zhenhui et al.}

\begin{abstract}
Retrieval systems primarily address the challenge of matching user queries with the most relevant advertisements, playing a crucial role in e-commerce search advertising. The diversity of user needs and expressions often produces massive long-tail queries that cannot be matched with merchant bidwords or product titles, which results in some advertisements not being recalled, ultimately harming user experience and search efficiency. Existing query rewriting research focuses on various methods such as query log mining, query-bidword vector matching, or generation-based rewriting. However, these methods often fail to simultaneously optimize the relevance and authenticity of the user's original query and rewrite and maximize the revenue potential of recalled ads. 

In this paper, we propose a \textbf{M}ulti-\textbf{o}bjective aligned \textbf{B}idword \textbf{G}eneration \textbf{M}odel (MoBGM), which is composed of a discriminator, generator, and preference alignment module, to address these challenges. To simultaneously improve the relevance and authenticity of the query and rewrite and maximize the platform revenue, we design a discriminator to optimize these key objectives. Using the feedback signal of the discriminator, we train a multi-objective aligned bidword generator that aims to maximize the combined effect of the three objectives. Extensive offline and online experiments show that our proposed algorithm significantly outperforms the state of the art. After deployment, the algorithm has created huge commercial value for the platform, further verifying its feasibility and robustness.
\end{abstract}

\begin{CCSXML}
<ccs2012>
   <concept>
       <concept_id>10002951.10003317.10003325.10003330</concept_id>
       <concept_desc>Information systems~Query reformulation</concept_desc>
       <concept_significance>500</concept_significance>
       </concept>
   <concept>
       <concept_id>10002951.10003317.10003325.10003329</concept_id>
       <concept_desc>Information systems~Query suggestion</concept_desc>
       <concept_significance>500</concept_significance>
       </concept>
   <concept>
       <concept_id>10010147.10010178.10010179.10010182</concept_id>
       <concept_desc>Computing methodologies~Natural language generation</concept_desc>
       <concept_significance>500</concept_significance>
       </concept>
   <concept>
       <concept_id>10002951.10003260.10003272.10003273</concept_id>
       <concept_desc>Information systems~Sponsored search advertising</concept_desc>
       <concept_significance>300</concept_significance>
       </concept>
 </ccs2012>
\end{CCSXML}

\ccsdesc[500]{Information systems~Query reformulation}
\ccsdesc[500]{Information systems~Query suggestion}
\ccsdesc[500]{Computing methodologies~Natural language generation}
\ccsdesc[300]{Information systems~Sponsored search advertising}

\keywords{
Large Language Model,
Query-bidword Generation,
Multi-objective Alignment,
E-Commerce Search Advertising
}



\maketitle

\section{INTRODUCTION}
For e-commerce search advertising, retrieval systems play a critical role in matching user queries with the most relevant advertisements. As online shopping becomes a lifestyle, the diversity of user needs and query expressions results in many long-tail queries that often elude effective matching with merchant bidwords or product titles. This challenge not only leads to suboptimal ad recall but also reduces search efficiency and negatively impacts user experience and platform revenue~\cite{yuan2023multi,yuan2024semi}. Consequently, enhancing the effectiveness of query rewriting mechanisms has become increasingly paramount.

Early research on query rewriting has explored a variety of methodologies, including mining-based methods~\cite{cui2002probabilistic,jones2006generating,antonellis2008simrank}, vector matching-based methods~\cite{chen2020rpm,li2022query}. While these approaches have shown good performance, they often fall short in optimizing three critical dimensions simultaneously: the relevance and authenticity of the user's original query to its rewrite and the potential revenue brought from recalled advertisements. For instance, data mining-based methods typically rely on user search session logs to derive query substitutions~\cite{jones2006generating} or user click data~\cite{antonellis2008simrank}, which can struggle with data scarcity, particularly for infrequently searched long-tail queries. Similarly, vector matching approaches~\cite{chen2020rpm,li2022query} focus on embedding similarities but can not fully leverage the revenue potential associated with ad placements.

Recent advancements in generation-based methods~\cite{wang2021queen,vakulenko2021question,ma2023query,zuo2023industry} have made strides in addressing the complexities of query rewriting. However, despite their enhanced capabilities, these methods are still hard to ensure that rewrites not only maintain relevance and authenticity but also maximize the commercial value of the ads presented. Furthermore, while large language models (LLMs) with reinforcement learning (RL) techniques~\cite{dai2024enhancing,peng2024large} have been employed to improve alignment with user preferences, they often face challenges related to training complexity and stability, particularly when dealing with multiple alignment objectives.

To address the above problems, this paper proposes a novel multi-objective aligned bidword generation model. MoBGM is composed of a generator, a discriminator, and a preference alignment module. We first post pre-train and fine-tune the generator utilizing e-commerce data. Subsequently, we design a discriminator to explicitly optimize three interrelated objectives: (1) query-bidword relevance, (2) generated bidword authenticity, and (3) advertising revenue. After obtaining the feedback signal from the discriminator, we design a multi-objective alignment module to further align MoBGM with human preference, for maximizing the combined effect of the three objectives. We conduct extensive experiments to evaluate the performance of MoBGM. The results from extensive offline and online experiments demonstrate that MoBGM significantly outperforms the existing state of the art, delivering substantial commercial value after deployment. This not only validates the feasibility and robustness of our approach but also underscores its potential to transform e-commerce search advertising practices. 

The contributions of this paper can be summarized as follows:
\begin{itemize} 
\item We propose a novel multi-objective aligned bidword generation model (MoBGM) that integrates three key objectives: query-bidword relevance, generated bidword authenticity, and advertising efficiency into a cohesive model. This model enhances the alignment between user queries and advertisement recall, addressing the limitations of existing methods.

\item We design an effective framework that leverages a reward signal derived from log data to simultaneously optimize for multiple objectives. Our model employs advanced techniques for training, allowing for more accurate and commercially viable query rewrites that cater to diverse user intents.

\item The effectiveness of MoBGM has been confirmed through extensive offline experiments on a large-scale real-world dataset, as well as online A/B testing. Furthermore, MoBGM has been deployed in production at an e-commerce platform, handling hundreds of millions of requests daily. This deployment not only highlights its practical applicability but also demonstrates its significant commercial value as a robust solution for large-scale query rewrite services. 
\end{itemize}

\section{RELATED WORK}
\subsection{Query Rewriting}
Query rewriting is a crucial module in e-commerce search systems, as it aligns user queries with the relevant advertisements, thereby significantly impacting both the user's shopping experience and the platform's revenue. The prevalent methods can be categorized into three types: data mining-based methods~\cite{cui2002probabilistic,jones2006generating}, vector matching-based methods~\cite{chen2020rpm,li2022query}, and generation-based methods~\cite{peng2024large,ma2023query}.

Data mining-based query rewriting methods typically extract query substitutions from user search session logs~\cite{jones2006generating} or construct graphs based on user click data~\cite{cui2002probabilistic,antonellis2008simrank}. These methods heavily depend on statistical features and user behavior data; however, they face challenges in dealing with data scarcity, particularly for infrequently searched long-tail queries.

Vector matching-based methods project the query and rewrite into a high-dimensional space, employing distance metrics (e.g., cosine similarity) to identify the closest rewrite embeddings~\cite{grbovic2015context,chen2020rpm}. Furthermore, \cite{li2022query} utilizes context and content logs to train semantically rich embeddings. While these techniques partially ensure relevance between queries and rewrites, they cannot maximize the revenue potential of the associated advertisements.

Generation-based methods leverage sequence-to-sequence generative models to create candidate rewrites for user input queries. Initial efforts by~\cite{riezler2010query,gao2012learning} employed statistical machine translation (SMT) frameworks, which were subsequently enhanced by deep learning approaches such as recurrent neural networks (RNN)\cite{sordoni2015hierarchical,he2016learning} and Transformers~\cite{vaswani2017attention}. Recent advancements, including attention-based models~\cite{wang2021queen,vakulenko2021question,zuo2023industry}, have further refined rewrite generation. With the development of large language models (LLMs), studies~\cite{peng2024large,ma2023query} have focused on fine-tuning LLMs to enhance the quality of generated rewrites. Despite their ability to address long-tail query rewriting issues, these methods still hardly maximize the relevance of queries to rewrites and the revenue they generate.

\subsection{Preference Alignment}
With the increasing number of parameters in pre-trained language models, these models exhibit emergent abilities~\cite{wei2022emergent}, as described by scaling laws~\cite{kaplan2020scaling}. However, they can also produce hallucinations and toxic content due to the nature of their training data and generation mechanisms. To mitigate these issues, \cite{instructGPT} employed RL techniques, such as Proximal Policy Optimization (PPO)~\cite{PPO,dai2024enhancing}, to fine-tune the generative model using human feedback. In the query rewriting scenarios, RL has also been applied to align the model for higher relevance~\cite{mohankumar2021diversity,agrawal2023enhancing,pang2025generative}. Despite their effectiveness, RL methods often entail complex training processes and lead to instability~\cite{casper2023open}.

Recognizing the limitations of RL, Direct Preference Optimization (DPO)~\cite{rafailov2024direct} derives a closed-form solution to the RL optimization objective and introduces a straightforward approach to model alignment using only a classification loss. Subsequently, methods like IPO~\cite{IPO}, KTO~\cite{KTO}, and SimPO~\cite{SimPO} modify the loss function terms in DPO to improve the performance of DPO. 

These preference alignment methods only focus on a single alignment objective, such as safety~\cite{llama2,qwen}. However, models usually need to be aligned with multiple objectives in real applications~\cite{MORLHF}. Simply using a weighted sum of multiple rewards for training is suboptimal. Multi-Objective direct preference optimization (MODPO)~\cite{MODPO} leverages margin reward models to balance the multi-objective rewards. Based on these findings, we propose a novel multi-objective alignment method to better align with three key objectives: relevance, authenticity, and advertising efficiency in the preference alignment stage.

\begin{figure*}[!htbp]
    \centering
    \includegraphics[scale=1.0]{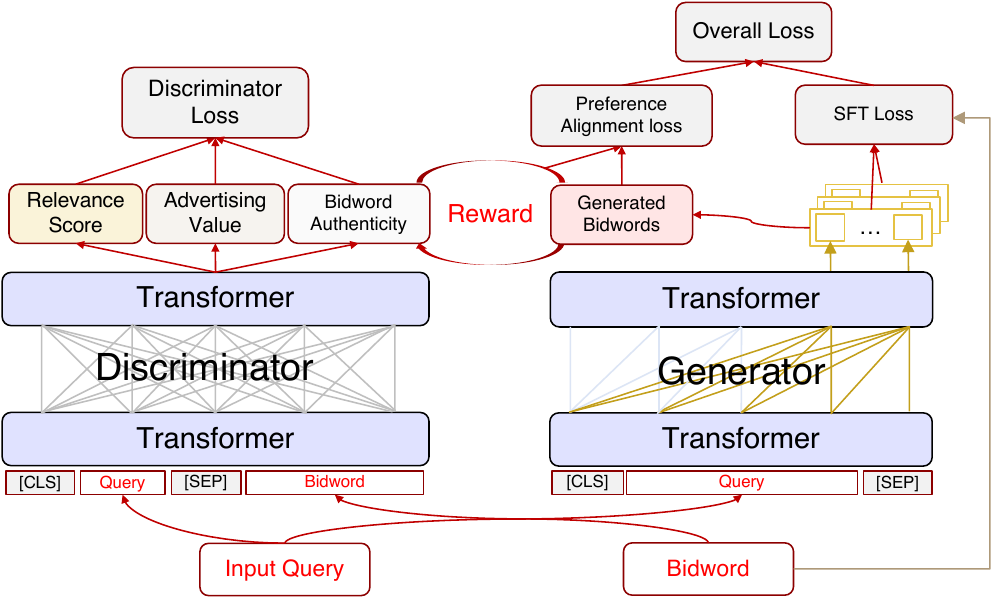}
    \caption{The architecture of the multi-objective aligned bidword generation model.}
    \label{model_structure}
\end{figure*}

\section{MODEL}
In this section, we first formally define the bidword generation task. Then, we introduce the modules of MoBGM in detail and analyze the model's influence during the training and prediction process.

\subsection{Problem Statement}
Suppose the query inputted by users on the E-commerce applications has $q = [x_1, x_2, \ldots, x_{L_q}]$ characters. Bidword rewriting reformulates the user's input query into a bidword that advertises buys to retrieve relevant products better. We aim to train a better rewrite model $M$ that can rewrite q as $q^{\prime} = M_{\theta}(q)$, where $q^{\prime}$ is the rewritten bidword that is used to retrieve products to complete the subsequent task.

The primary challenges are evaluating the value of the rewritten bidword and guiding the model to generate high-value bidwords that enhance the efficiency of recall and overall ad revenue.

\subsection{Overview}
Figure~\ref{model_structure} illustrates the components of the MoBGM, which consists primarily of two modules: the discriminator and the generator. Specifically, the discriminator treats the query and the generated bidword as inputs, computing three key metrics: the relevance score, the authenticity score, and the advertising value of the generated bidword. These predicted scores are utilized to assess the generated bidword's overall value and inform the model's training process. The generator, on the other hand, takes the query as input and produces a set of bidwords that are pertinent to the user's query. The interaction between the generator and discriminator is designed to enhance the overall value of the generated bidword.

\subsection{Generator Training}

\subsubsection{Post Pre-training}
LLMs have demonstrated exceptional performance on general natural language processing tasks, primarily benefiting from the extensive knowledge of natural language expressions acquired during their pre-training phase. However, in e-commerce applications, the distribution of user input queries and product data significantly differs from that of natural language expressions. Open-source LLMs without post-training, also exhibit mediocre performance on e-commerce natural language processing tasks; a similar phenomenon has been observed with classification-based models, such as BERT~\cite{kenton2019bert} and RoBERTa~\cite{liu1907roberta}.

Therefore, it is necessary to conduct post-training open-source LLMs using e-commerce corpora to enable the models to learn the specific knowledge and common expression habits prevalent in e-commerce text. We constructed a training dataset $D_\mathrm{PPT}$ based on query-product click behavior. By selecting a batch of query-product title data sorted by click frequency, we developed two pre-training sub-tasks: query to the product title, and product title to query. The training objective can be formally defined as follows: 
\begin{equation}
\begin{split}
\mathcal{L}_\mathrm{PPT}(\pi_{\theta}) = & - \mathbb{E}_{(q, t) \sim \mathcal{D}_\mathrm{PPT}} \Bigg( \sum_{i=1}^{L_q} \log \pi_{\theta}(q_i | q_{<i}, t) + \sum_{j=1}^{L_t} \log \pi_{\theta}(t_j | t_{<j}, q) \Bigg) \,, 
\end{split} 
\end{equation}
where $L_q$ and $L_t$ denote the length of the query and product title, respectively.

This approach enables the large language model to comprehensively learn the characteristics of user input queries and the knowledge embedded in product titles within the e-commerce context, thereby enhancing the model's suitability for e-commerce tasks.

\subsubsection{Supervised Fine-tuning}
Pre-training with the domain data enhances the natural language understanding and generation capabilities of LLMs in the e-commerce domain. To further improve their performance in e-commerce query rewriting, we introduce task-specific supervised fine-tuning (SFT). The process of generating text with a conditional language model can be conceptualized as a constrained auto-regressive sampling strategy. Given a query $q$ and its corresponding gold standard $b$, the training objective is to maximize the conditional probability $Pr(b|q)$. Specifically, the training objective for the rewriting model involves minimizing the negative log-likelihood: 
\begin{equation}
\begin{split}
& \mathcal{L}_{SFT}(\pi_{\theta}) = - \mathbb{E}_{(q, b) \sim \mathcal{D}_\mathrm{SFT}} \sum_{i=1}^{L_b} \log \pi_\theta(b_i | q, b_{<i})  \,, \\
\end{split}
\end{equation}
where $b_i$ is the $i$-th token of the bidword $b$, $\mathcal{D}_\mathrm{SFT}$ is the collected training data, $L_b$ is the length of the bidword.

\subsection{Discriminator Training}  \label{discriminator_section}
\subsubsection{Query-bidword relevance objective}
Ensuring query and bidword relevance is the first step to retrieving the users' desired products. To maintain the user search experience and improve the system's advertising distribution efficiency, we need to train a relevance discriminator to evaluate the relevance of the query and generated bidword. 

We define four types of rewrite relations: synonym, hypernym paraphrase to hyponym, hyponym paraphrase to hypernym, and incorrect paraphrase. To train a relevance model, we collect query and bidword pair data from the online search log and give them to experts to annotate the four types of rewrite relations. The data with consistent annotation results from two experts are treated as reliable data to construct training sets and validation sets.

Suppose the query token sequence is $q = [q_1, q_2, \ldots, q_{L_q}]$ and the bidword token sequence is $b = [b_1, b_2, \ldots, b_{L_b}]$. The relevance predictor of query and bidword is defined as follows:
\begin{equation}
\begin{split}
& \mathbf{X}_i = \mathrm{BERT_{[CLS]}}\left(\left[q_1, q_2, \ldots, q_{L_q}, \mathrm{[SEP]}, b_1, b_2, \ldots, b_{L_b} \right] \right)   \,, \\
& \mathbf{Pr_{rele}}(q, b) = \mathbf{Softmax} \left(\mathbf{X}_i \mathbf{W}^T + \mathbf{b} \right) \,, \\
& \mathcal{L}_{rele} = - \sum ^{|C|}_{i=1} \mathbf{y}_i \log \left( \mathbf{Pr_{rele}}(q, b) \right) \,, 
\end{split}
\end{equation}
where $ \mathrm{BERT_{CLS}}$ is the "[CLS]" representation of the last layer of BERT. $|C|$ denotes the total number of classes. $\mathbf{X}_i \in \mathbb{R}^{1 \times d}$ and $\mathbf{W} \in \mathbb{R}^{|C| \times d} $  denote the token embedding matrix of input and category, respectively.

\subsubsection{Bidword authenticity objective}
It is crucial to note that e-commerce query rewriting is different from other text generation tasks. In this scenario, achieving semantic similarity between the original query and its rewritten version does not automatically guarantee the retrieval of a similar set of products. If a user inputs the query \textit{"washing machine"} and the model generates a rewrite like \textit{"4K 144Hz washing machine"}, it might achieve high relevance, but it might fail to retrieve any products due to its low authenticity. We aim to ensure a robust correlation between the products obtained through the rewritten query and those related to the original query. Therefore, beyond ensuring semantic relevance, the rewritten query must be formulated as a bidword; this is essential to enable the retrieval of advertisements associated with that bidword.

After checking the outputs of model generation, we observed that some generated queries satisfy the relevance, yet deviate from established user search intention. These synthesized queries either fail to align with the advertiser's bidwords or exhibit long-tailed distribution characteristics, resulting in diminished recall efficiency. Previous research~\cite {lian2019end,song2021triangular,agrawal2023enhancing} has employed prefix tree (Trie) constraints to guarantee that all generated reconstructions adhere to a rigorously defined set. However, our empirical findings indicate that the employment of the Trie decoding technique could compromise the relevance of the rewritten query to the original query. Consequently, we have developed a model capable of discerning whether a generated query represents one a user is likely to search for.

We collect user query data from search logs and sort them in reverse order of search frequency. When a query can match a bidword and the search frequency exceeds a threshold, the query is considered a positive sample; otherwise, it is marked as a negative sample. The authenticity of the generated query is defined as a binary classification task. Suppose that the bidword token sequence is $b = [b_1, b_2, \ldots, b_{L_b}]$. The authenticity predictor of bidword is defined as follows:
\begin{equation}
\begin{split}
& \mathbf{X}_i = \mathrm{BERT_{[CLS]}} \left(\left[b_1, b_2, \ldots, b_{L_b} \right] \right)   \,, \\
& \mathbf{Pr}_{au}(b) = \mathbf{Softmax}(\mathbf{X}_i\mathbf{W}^T + \mathbf{b}) \,, \\
& \mathcal{L}_{au} = - \sum \mathbf{y} \log \left( \mathbf{Pr}_{au}(b)  \right) + \left( 1 - \mathbf{y} \right) \log \left( 1 - \mathbf{Pr}_{au}(b) \right) \,,
\end{split}
\end{equation}
where $ \mathrm{BERT_{CLS}}$ is the "[CLS]" representation of the last layer of BERT. $\mathbf{y}$ is a binary value (0 or 1), indicating whether the bidword is authenticated or not. $\mathbf{X}_i \in \mathbb{R}^{1 \times d}$ and $\mathbf{W} \in \mathbb{R}^{|C| \times d} $  denote the token embedding matrix of input and category, respectively.

\subsubsection{Advertising value}
The relevance objective and authenticity objective can ensure the retrieval efficiency and relevance between query and retrieved products to some extent, bringing a good experience for users and advertisers. However, high retrieval efficiency and relevance may not necessarily bring growth in advertising revenue for e-commerce platforms. Therefore, we use the platform’s advertising distribution efficiency indicator as a kind of alignment preference. 

We use the CPM (Cost per Mille), which is a commonly used metric in advertising systems, as an optimization objective. We collect bidword data from the online click logs and compute the CPM value for every bidword. The target is to make the model fit the CPM value $\mathrm{CPM}_{b}$ of the bidword $b = [b_1, b_2, \ldots, b_{L_b}]$, thus, we use the Mean Squared Error Loss as the training objective. The process can be formulated as follows:
\begin{equation}
\begin{split}
& \mathbf{X}_i = \mathrm{BERT_{[CLS]}} \left(\left[b_1, b_2, \ldots, b_{L_b} \right] \right)   \,, \\
& \hat{y}_{val}(b) = \mathbf{W}^T\mathbf{X}_i  \,, \\
& \mathcal{L}_{val} = \Vert \hat{y}_{val}(b) - \mathrm{CPM}_{b} \Vert^2 \,, \\
\end{split}
\end{equation}
where $\mathbf{W} \in \mathbb{R}^{1 \times d} $ is a transformation matrix.

\subsection{Multi-Objective Preference Alignment}

\subsubsection{Multi-Objective Preference Reward}
After the generator and discriminator training, we leverage the labels assigned by the discriminator as the preference reward further to finetune the LLM so as to align with the primary objective of bidword generation. Specifically, the preference reward encompasses three dimensions: relevance, authenticity, and advertising value, which are elaborated upon below. 

The query-bidword relevance reward can be defined as follows: 
\begin{equation}
\begin{split}
& \mathbf{R}_{rele}(q, b) = \mathbf{Pr}_{syn}(q, b) + \mathbf{Pr}_{hyper}(q, b) - \mathbf{Pr}_{error}(q, b)\,, \\
\end{split}
\end{equation}
where $\mathbf{Pr}_{syn}(\cdot)$, $\mathbf{Pr}_{hyper}(\cdot)$, and $\mathbf{Pr}_{error}(\cdot)$ denote the probability that the relation between query and bidword is predicted to synonym, a hypernym, and an error rewrite. For a given query $q$, we select the bidword $b_w$ as a winner if it has a higher $\mathbf{R}_{rele}(q, b_w)$, and $b_l$ as the loser if it has a lower $\mathbf{R}_{rele}(q, b_l)$.

The bidword authenticity reward can be defined as follows: 
\begin{equation}
\begin{split}
& \mathbf{R}_{au}(b) = \mathbf{Pr}_{au}(b) \,, \\
& \Delta \mathbf{R}_{au} = \mathbf{R}_{au}(b_w) - \mathbf{R}_{au}(b_l)  \,,  \\
\end{split}
\end{equation}
where $\mathbf{Pr}_{au}(b)$ is the probability that the generated bidword is predicted as an authentic bidword.

The advertising value reward of the generated bidword can be defined as follows: 
\begin{equation}
\begin{split}
& \mathbf{R}_{val}(b) = \hat{y}_{val}(b) \,, \\
& \Delta \mathbf{R}_{val} = \mathbf{R}_{val}(b_w) - \mathbf{R}_{val}(b_l)  \,,  \\
\end{split}
\end{equation}
where $\hat{y}_{val}(b)$ is the estimated CPM value of the generated bidword.

\subsubsection{Preference Alignment Loss}
To accommodate the diversity of human preferences and the intricacies involved in optimizing multiple objectives with reinforcement learning (RL), we extend MODPO~\cite{MODPO}, a stable and efficient extension of DPO that precisely optimizes during the Preference Alignment (PA) process as described in Eq.~\ref{eq:modpo} without the need for RL.
\begin{flalign}
\label{eq:modpo}
\mathcal{L_\mathrm{PA}} (\pi_{\theta}) = & -\mathbb{E}_{(q, b_w, b_l) \sim \mathcal{D}_\mathrm{PA}} \left[ \log\sigma\left( \frac{\beta_{w}}{\mathbf{w}_{rel}}\log\frac{\pi_{\theta}(b_w | q)}{\pi_\mathrm{SFT}(b_w | q)} \right. \right. & \nonumber \\ 
& \left. \left. 
- \frac{\beta_l}{\mathbf{w}_{rel}}\log\frac{\pi_{\theta}(b_l | q)}{\pi_\mathrm{SFT}(b_l | q)} - \frac{\mathbf{w}_{au}}{\mathbf{w}_{rel}}\Delta\mathbf{R}_{au} - \frac{\mathbf{w}_{val}}{\mathbf{w}_{rel}}\Delta\mathbf{R}_{val} \right) \right] &
\end{flalign}
where $\pi_\mathrm{SFT}$ is fixed as a reference model for the KL-divergence penalty, and $\beta_w$, $\beta_l$ control the KL penalty of the winner and loser. $\pi_{\theta}$ is initialized from $\pi_\mathrm{SFT}$ and is trained during the preference alignment process. $\mathbf{w}_{rel}$, $\mathbf{w}_{au}$, and $\mathbf{w}_{val}$ represent the weight of the relevance, authenticity, and advertising value, respectively. This multi-objective approach enables the flexible customization of language models to accommodate diverse preference distributions through adjusted reward weightings during fine-tuning.

\begin{table*}[!htbp]
    \centering
    \caption{Dataset statistics.}
    \setlength{\tabcolsep}{5mm}{
        \begin{tabular}{c|cc|cc|c}
            \toprule
            \multirow{2}{*}{\textbf{Statistics}} & \multicolumn{2}{c|}{\textbf{SFT Dataset}} & \multicolumn{2}{c|}{\textbf{Preference Alignment Dataset}} & \multirow{2}{*}{\textbf{Golden Dataset}}  \\ 
            &\textbf{Train} &\textbf{Validation} &\textbf{Train} &\textbf{Validation} &  \\   
            \hline
            \hline 
            \#Uniq. Queries  & 17,778,056 & 98,543 & 984,271 & 49,384 & 90,000 \\
            \hline
            \#Uniq. Bidwords & 706,702 & 77,382 & 223,680 & 20,335 & 125,909 \\
            \hline
            \#Query-Bidword Pairs & 37,253,452 & 100,000 & 7,466,588 & 100,000 & 900,000 \\
            \hline
            \#Bidwords per query & 2.10 & 1.01 & 7.59 & 2.02 & 10.00 \\
            \hline
            Avg. chars & 5.86 & 5.84 & 4.12 & 4.13 & 4.54 \\
            \hline
            \bottomrule
        \end{tabular}
    }
    \label{tab:datset}
\end{table*}

\begin{table*}[!htbp]
  \caption{
    The experimental results are compared to the baselines on the Golden Dataset. 
  }
  \centering
  \label{main_experiment}
  \setlength{\tabcolsep}{2mm}{
      \begin{tabular}{c|cccccccc}
            \toprule
            \multirow{2}{*}{\textbf{Models}}  & 
            \multicolumn{8}{c}{\textbf{Golden Dataset (\%)}} \\
            &\textbf{Recall@3} &\textbf{Recall@5} &\textbf{Recall@10}  &\textbf{NDCG@3} &\textbf{NDCG@5} &\textbf{NDCG@10} & \textbf{Relevance} & \textbf{Authenticity}  \\
            \midrule
            \midrule
            GQS~\cite{jones2006generating}            & 5.24 & 7.31  & 10.63  & 12.21  & 10.82  & 9.82  & 21.31  & \textbf{98.86} \\
            Simrank++~\cite{antonellis2008simrank}            & 6.25 & 9.16 & 13.30  & 14.06  & 12.95  & 10.64  & 59.69  & 98.13 \\
            RQRF~\cite{chen2020rpm}            & 8.44 & 13.70  & 23.57  & 18.72  & 18.50  & 17.17  & 97.45  & 67.21 \\ 
            \midrule
            EGRM~\cite{lian2019end}  & 12.18 & 20.14  & 38.49 & 26.97  & 27.03  & 26.90  & 72.17  & 95.43 \\
            BART~\cite{bart}            & 11.20 & 19.92  & 39.67 & 26.53  & 26.66  & 27.27  & 80.10  & 93.69 \\
            CLOVER~\cite{mohankumar2021diversity}       & 10.78 & 17.95  & 35.78  & 23.73  & 23.93  & 24.52  & 91.72  & 87.48 \\
            EEQR~\cite{dai2024enhancing}       & 14.60 & 21.73  & 39.28  & 32.10  & 30.38  & 28.74  & 90.17  & 94.56 \\
            BEQUE~\cite{peng2024large}  & 13.30 & 22.13  & 39.87  & 29.42  & 29.66  & 28.52  & 92.58  & 91.12 \\
            \midrule
            \textbf{MoBGM}   & \textbf{15.18} & \textbf{22.47}  & \textbf{40.67}  & \textbf{33.38}  & \textbf{31.06}  & \textbf{29.42}  & \textbf{97.49}  & 98.65 \\
            \bottomrule
        \end{tabular}
    }
\end{table*}

\section{EXPERIMENT}
This section will discuss the offline and online experiments in detail. We first introduce the datasets and the evaluation metrics used in this paper. Then, we analyze the experiment results by several fair comparisons with strong baselines. After that, we deeply investigate the effect of different modules of the MoBGM. Subsequently, we present the online performance of the model on the JD search engine and further analyze the influence of various modules.

\subsection{Dataset}
\label{sec:Dataset}
To train the MoBGM and evaluate its effectiveness and generality, we conducted extensive experiments on three large-scale real-world datasets collected from users' search and click logs on the JD application. The statistics of the datasets are listed in Table~\ref{tab:datset}. Specifically, 
\begin{itemize}
    \item \textbf{Pretraining Dataset}: For each product in the advertising systems, we selected the corresponding high-frequency queries in the online logs to construct product-query pairs by users' click behavior. In total, we obtained approximately 300 million samples for the pretraining stage. 
    
    \item \textbf{SFT Dataset}: Through mining from session logs, click graphs, and semantic matching, we identified approximately 6 billion query-bidword pairs. We applied down-sampling to the high-frequency queries and up-sampling to those long-tail queries. Furthermore, we calculated the relevance, authenticity, and advertising efficiency of the bidword. We then filtered out query-bidword pairs that fell below the threshold, resulting in an SFT dataset of approximately 37 million query-bidword pairs.

    \item \textbf{Preference Alignment Dataset}: For user queries, we categorized them into head, middle, and tail categories based on the frequency in the online logs. From each category, we uniformly sampled 360,000 queries, with 330,000 designated for the preference alignment process and 30,000 reserved for the golden evaluation set. Consequently, the total number of queries in the preference alignment dataset is 990,000.

    \item \textbf{Golden Dataset}: For the sampled 90000 queries distinct from the Preference Alignment Dataset, we use a heuristic policy to identify the 10 most valuable bidwords from the click logs. This set will be used as the golden standard to evaluate performance.
\end{itemize}

\subsection{Baseline Models}
To evaluate the efficacy of our proposed method, we compare MoBGM with several strong baseline models, including data mining-based methods, matching-based methods, and generation-based methods. The detailed introductions are listed as follows: 

\noindent (1) Data mining-based methods and matching-based methods: 
\begin{itemize}
    \item\textbf{GQS}~\cite{jones2006generating}: It utilizes user-session logs to mine rewrite candidates and selects high-quality ones based on predefined criteria.
    \item \textbf{Simrank++}~\cite{antonellis2008simrank}: It exploits the click graph structure and introduces the weights of the edges in the click graph supporting the similarity between queries. 
    \item \textbf{RQRF}~\cite{chen2020rpm}: It embeds both queries and bidwords to vectors in the same implicit space, converting the rewriting probability between each query and bidword to the distance between the two vectors. 
\end{itemize}

\noindent (2) Generation-based methods: 
\begin{itemize}
    \item \textbf{EGRM}~\cite{lian2019end}: It is an LSTM-based generative model with the Trie decoding method for bidword generation. 
    \item \textbf{BART}~\cite{bart}: It is an encoder-decoder structured generative model previously deployed in JD's advertising system. We train the BART model with the same settings as we conducted for LLM. 
    \item \textbf{CLOVER}~\cite{mohankumar2021diversity}: It proposes a diversity-driven RL-based algorithm to generate both high-quality and diverse reformulations by optimizing for human assessment of rewrite quality.
    \item \textbf{EEQR}~\cite{dai2024enhancing}: It utilizes a PPO loss and a standard SFT loss during the preference alignment process.
    \item \textbf{BEQUE}~\cite{peng2024large}: It utilizes the PRO method during the preference alignment stage, which involves a ranking loss and an SFT loss. 
\end{itemize}

\subsection{Evaluation Metrics}
Following the methodologies employed in prior research~\cite{li2022query,peng2024large}, we utilize Recall@k and NDCG@k as our primary metrics for performance evaluation. Additionally, we compute average relevance and authenticity as supplementary metrics to facilitate a more nuanced assessment. The definitions of these metrics are as follows:
\begin{itemize}
\item \textbf{Recall@k}: The number of top $k$ generated results that are included in the standard set, to the total number of items in the standard set.
\item \textbf{NDCG@k}: The normalized discounted cumulative gain, a metric designed to evaluate ranking quality by taking into account the positions of relevant items within the generated result list.
\item \textbf{Relevance / Authenticity}: The percentage of generated bidwords that successfully pass relevance or authenticity filters, as assessed by human experts.
\end{itemize}

\subsection{Experiment Settings}
We implemented our models using the PyTorch framework and utilized the Llama-3-Chinese-8B-Instruct~\cite{chinese-llama-alpaca} model as the foundational large language model (LLM). For optimization, we employed the AdamW optimizer~\cite{adamw2019decoupledweightdecayregularization}. During the pre-training phase, the model was trained for one epoch, followed by two epochs in the SFT stage, both with a learning rate of 1e-5. In the preference alignment phase, the model underwent training for two epochs with a learning rate of 1e-6. The model training incorporated a warm start strategy with a cosine decay schedule. For bidword generation, we utilized the beam search algorithm with a length penalty of 2.0 and a maximum of 16 new tokens. All training was conducted on 8 H-800 GPUs, with a batch size of 64 per GPU.

We set $\beta_w=1.0$ and  $\beta_l=0.25$ in the experiments. Additionally, we set $\mathbf{w}_\mathrm{rel}=0.5$, $\mathbf{w}_\mathrm{au}=0.2$, and $\mathbf{w}_\mathrm{val}=0.3$. We select the optimal parameter configuration based on the performance on the validation set and then evaluate this configuration on the golden set.

\begin{table*}[!htbp]
\caption{Ablation study of the proposed model MoBGM on the Golden Dataset.}
  \centering
  \label{ablation_study}
  \setlength{\tabcolsep}{1.5mm}{
      \begin{tabular}{c|cccccccc}
            \toprule
            \multirow{2}{*}{\textbf{Models}}  & 
            \multicolumn{8}{c}{\textbf{Golden Dataset (\%)}} \\
            &\textbf{Recall@3} &\textbf{Recall@5} &\textbf{Recall@10}  &\textbf{NDCG@3} &\textbf{NDCG@5} &\textbf{NDCG@10} & \textbf{Relevance} & \textbf{Authenticity}  \\
            \midrule
            
            \textbf{MoBGM}   & \textbf{15.18} & \textbf{22.47}  & 40.67  & \textbf{33.38}  & \textbf{31.06}  & \textbf{29.42}  & 92.49  & 98.65 \\
            w. Trie decoding & 12.25 (-2.93) & 21.11 (-1.36) & \textbf{40.92} (+0.25) & 26.99 (-6.39) & 27.29 (-3.77) & 28.04 (-1.38) & 87.25 (-5.24) & \textbf{98.82} (+0.17) \\
            w/o. finetune    & 12.28 (-2.90) & 20.44 (-2.03) & 40.55 (-0.12) & 27.03 (-6.35) & 27.23 (-3.83) & 27.92 (-1.50) & 92.25 (-0.24) & 96.88 (-1.77) \\
            w/o. alignment   & 14.17 (-1.01) & 22.10 (-0.37) & 38.16 (-2.51) & 30.15 (-3.23) & 30.10 (-0.96) & 27.76 (-1.66) & 88.71 (-3.78) & 92.92 (-5.73) \\
            w/o. multi obj.  & 14.59 (-0.59) & 22.39 (-0.08) & 38.88 (-1.79) & 32.05 (-1.33) & 30.92 (-0.14) & 28.36 (-1.06) & 88.23 (-4.26) & 94.90 (-3.75) \\ 
            w/o. relevance   & 12.54 (-2.64) & 19.15 (-3.32) & 36.42 (-4.25) & 27.72 (-5.66) & 27.45 (-3.61) & 27.91 (-1.51) & 75.24 (-17.25) & 98.71 (+0.06) \\
            w/o. authenticity& 14.56 (-0.62) & 22.01 (-0.46) & 40.12 (-0.55) & 32.97 (-0.41) & 30.88 (-0.18) & 29.16 (-0.26) & 93.31 (+0.82) & 90.85 (-7.80) \\
            w/o. CPM         & 14.24 (-0.94) & 21.23 (-1.24) & 39.56 (-1.11) & 31.24 (-2.14) & 29.49 (-1.57) & 27.28 (-2.14) & \textbf{93.47} (+0.98) & 98.75 (+0.10) \\
            \bottomrule
        \end{tabular}
    }
\end{table*}

\subsection{Offline Evaluation}

\subsubsection{Offline performance}
The experimental results on the preference alignment dataset are reported in Table~\ref{main_experiment}. Overall, the experimental results indicate that MoBGM significantly outperforms all baselines on the real-world dataset. Specifically, we have the following observations:

(1) For data mining-based methods and matching-based methods (i.e., GQS, Simrank++, RQRF), it is obvious that MoBGM outperforms them by a significant margin on the dataset. Data mining-based methods mainly focus on statistical features and user behavior data, such as search sessions or click behaviors. It performs well for frequently searched queries. However, these methods are powerless for long-tail queries, because these queries lack user interaction data, thus it is difficult to mine the rewriting result of the candidate. For the matching-base method, they can partially ensure relevance between queries and rewrites. However, these methods are difficult to ensure that recalled rewrites can retrieve products and bring an increase in advertising revenue. 

(2) Compared with recently proposed generation-based query rewrite methods (i.e., EGRM, BART, CLOVER, EEQR, BEQUE), MoBGM also demonstrates superior performance on most metrics of the Golden Dataset. As the results are shown in the table, Recall@k and NDCG@k obtain more than 1\% absolute improvement compared with the state-of-the-art. Moreover, the relevance and authenticity have also been improved to varying degrees. For previous generation-based methods, despite their ability to address long-tail query rewriting issues, they cannot ensure the authenticity of the generated query and its relevance with the original query, and at the same time, it is impossible to maximize the advertising revenue brought by the generated query. This paper addresses this problem by leveraging reward signals derived from the discriminator and designing a multi-objective alignment module to integrate three key objectives into a cohesive model.

In conclusion, MoBGM demonstrates a substantial improvement over all baseline models in terms of Recall@k, NDCG@k scores, relevance, and authenticity. The results confirm that the feedback signal from the discriminator effectively facilitates the simultaneous optimization of both the relevance and authenticity of the original query and its rewrites, while also maximizing the revenue potential of the recalled advertisements.

\subsubsection{Ablation study}
To further figure out the relative importance of each module in the proposed model, we perform a series of ablation studies over the different components of MoBGM. The variants of MoBGM are listed below: 
\begin{itemize}
    \item \textbf{w. Trie decoding}: Employing constrained Trie decoding during generation. 
    
    \item \textbf{w/o. finetune}: Removing the fine-tune steps and directly using post pre-trained MoBGM to conduct multi-objective preference alignment.
    
    \item \textbf{w/o. alignment}: A base LLM model that undergoes post-pretraining and supervised fine-tuning, without incorporating a preference alignment process.

    \item \textbf{w/o. multi obj.}: Removing the multi-objective alignment algorithm and instead applying a basic DPO method in the preference alignment process.
    
    \item \textbf{w/o. relevance}: Removing the relevance alignment when conducting multi-objective preference alignment.
    
    \item \textbf{w/o. authenticity}: Removing the authenticity alignment when conducting multi-objective preference alignment.
    
    \item \textbf{w/o. CPM}: Removing the CPM alignment when conducting multi-objective preference alignment. 
\end{itemize}

The experimental results are shown in Table~\ref{ablation_study}. We can observe that:

(1) When constrained Trie decoding is employed during generation, relevance tends to decrease, leading to a lower recall of items within the top 3 and top 5 results. When considering the top 10 results, recall may surpass the baseline due to improved authenticity, but the NDCG is reduced, indicating a suboptimal ranking order.

(2) Removing fine-tuning, the relevance, and authenticity still align with baseline levels, but both recall and NDCG fall below the baseline due to the reduced capability in generating bidwords.

(3) Eliminating alignment or utilizing a naive DPO method for alignment leads to a decrease in both relevance and authenticity compared to the baseline. Consequently, recall and NDCG are also reduced to varying extents.

(4) Finally, when relevance, authenticity, or CPM are individually removed, there is a significant drop in the corresponding metric, while other semantic metrics increase. Nevertheless, both recall and NDCG decline. Among these, relevance is the most crucial, whereas authenticity is the least important. This observation suggests that assigning a higher weight to relevance and a lower weight to authenticity could be beneficial.

\subsection{Online Evaluation}

\subsubsection{Online Deployment}
To reduce the response latency of online deployment, we first distill the parameters of MoBGM from the 8B base to 1B. Then, we utilized the MoBGM to pre-generate 15 bidwords for each of the top 10 million high-frequency queries from the online logs, storing them as an online cache. For queries ranked beyond the top 10 million, we use distilled MoBGM to generate 15 bidwords in real-time. In this way, the MoBGM can serve all traffic without increasing the latency of the entire search.

\begin{figure}[t]
\centering
\includegraphics[scale=0.63]{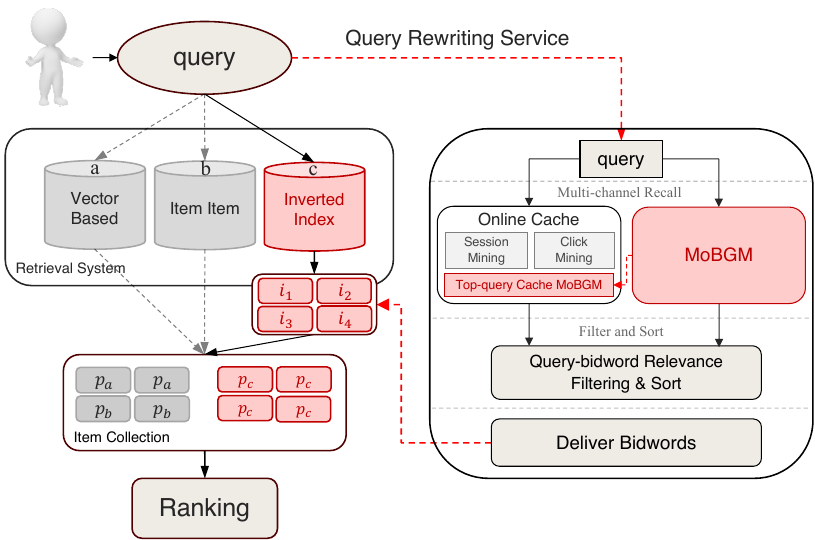}
\caption{The deployment of MoBGM and the role of the query rewrite module play in the E-commerce search system.}
\label{fig:system}
% \vspace{-1\baselineskip} 
\end{figure}

Figure~\ref{fig:system} illustrates the role the MoBGM plays in the JD search system. When a user submits a query, the system invokes the query rewrite service. Initially, it searches for the query in the online cache, which encompasses multiple channels of rewrite candidates, including bidwords generated by the MoBGM. If the user's query is not present in the online cache, the query rewriting service will call the MoBGM to generate 15 bidwords in real time. Once the rewrites are obtained, they are forwarded to the query-bidword relevance filtering and sorting module, which filters out irrelevant rewrites. Subsequently, the bidwords that pass this filtering process are sent to the inverted-index retrieval module to retrieve products. The retrieved items are then integrated with items from other retrieval branches and subjected to filtering by a sub-module designed to exclude irrelevant items that do not align with the user's intent. Finally, the filtered items are forwarded to the ranking module.

Overall, the bidwords generated by the MoBGM primarily influence the inverted-index retrieval branch during the product retrieval process. This mechanism aids in retrieving a greater number of products that align with user demands, thereby enhancing user experience and the efficiency of the search system.

\subsubsection{Online Performance}
Before being launched in production, we routinely deployed the MoBGM online on the JD search engine and made it randomly serve 5\% traffic as the test group. For a fair comparison, we also built a base group that uses the previous online model, serving 5\% traffic. During the A/B testing period, we monitor the performance of MoBGM and compare it with the online model. This period lasts for at least one week. 

For online evaluation, we use some business metrics: Ad. impressions(total count of ad impressions), CPC, CPM, and Ad. revenue(total ad revenue). The specific computations of CPC and CPM are defined as follows: 
\begin{equation}
\begin{split}
    & CPC = \frac{Ad. revenue}{\#Clicks}  \,, \\
    & CPM = 1000*\frac{Ad. revenue}{\#Impressions}    \,,  \\
\end{split}
\end{equation}
where $\#Clicks$ represents the total number of user clicks on an advertisement, while $\#Impressions$ denotes the total number of times the advertisement is displayed to users.

\begin{table}[!htbp]
  \centering
  \caption{Online improvements of the MoBGM. Improvements are statistically significant with $p < 0.05$  on a paired t-test. All performances of MoBGM and its variants are compared with the online model.}
  \label{online_performance}
  \setlength{\tabcolsep}{1mm}{
      \begin{tabular}{c|cccc}
            \toprule
            Models &\textbf{Ad. impressions}  &\textbf{CPC} &\textbf{CPM}  & \textbf{Ad. revenue}
            \\
            \midrule
            Online          & - & -  & - & -   \\
            MoBGM           & +0.39\%   & +1.72\%  & +1.74\%    & +2.13\%   \\
            w/o. finetune   & +0.13\%   & +1.46\%  & +1.42\%    & +1.55\%   \\
            w/o. multi-obj. & +0.26\%   & +0.26\%  & +0.32\%    & +0.58\%   \\
            \bottomrule
        \end{tabular}
    }
\end{table}

The online A/B experimental results are shown in Table~\ref{online_performance}. We can observe that the MoBGM consistently outperforms its variant models. (1) fine-tuning is the primary driver of the increase in advertising impressions, more impressions will bring more clicks and ad. revenue; (2) multi-objective alignment significantly enhances the CPC, which deep reason is its rewrites retrieve more relevant products and makes the bidding process more intense, thereby leads to an increase in CPM (cost per mile) and ad. revenue; (3) the simultaneous growth in advertising impressions and CPM contributes to the overall increase in advertising revenue. In conclusion, both the results of the offline evaluation and online A/B experiments consistently show the effectiveness and efficiency of the MoBGM.

\section{CONCLUSION AND FUTURE WORK}
In conclusion, this paper presents a significant advancement in the field of e-commerce search advertising through the introduction of the multi-objective aligned bidword generation model (MoBGM). By effectively integrating the critical dimensions of query-bidword relevance, generated bidword authenticity, and advertising efficiency, MoBGM addresses the prevalent challenges faced by existing query rewriting methodologies. Our comprehensive approach not only enhances the alignment between user queries and relevant advertisements but also contributes to improved user satisfaction and increased revenue for e-commerce platforms. The extensive offline and online experiments conducted validate the robustness and effectiveness of our proposed model, demonstrating its superiority over state-of-the-art methods. The successful deployment of MoBGM in a high-traffic e-commerce environment further underscores its practical applicability and commercial viability, handling hundreds of millions of requests daily. This substantial real-world impact signifies a transformative potential for query rewriting practices in e-commerce search advertising.

Future research could focus on enhancing the MoBGM framework by incorporating additional dimensions of user behavior and preferences. By integrating user-specific information, the model can develop personalized rewriting capabilities, improving both the user experience and the efficiency of product distribution.

\clearpage

\bibliographystyle{ACM-Reference-Format}
\balance
\bibliography{sample-base}


\begin{thebibliography}{42}

%%% the Web address.

\ifx \showCODEN    \undefined \def \showCODEN     #1{\unskip}     \fi
\ifx \showDOI      \undefined \def \showDOI       #1{#1}\fi
\ifx \showISBNx    \undefined \def \showISBNx     #1{\unskip}     \fi
\ifx \showISBNxiii \undefined \def \showISBNxiii  #1{\unskip}     \fi
\ifx \showISSN     \undefined \def \showISSN      #1{\unskip}     \fi
\ifx \showLCCN     \undefined \def \showLCCN      #1{\unskip}     \fi
\ifx \shownote     \undefined \def \shownote      #1{#1}          \fi
\ifx \showarticletitle \undefined \def \showarticletitle #1{#1}   \fi
\ifx \showURL      \undefined \def \showURL       {\relax}        \fi
% invisible to TeX
\providecommand\bibfield[2]{#2}
\providecommand\bibinfo[2]{#2}
\providecommand\natexlab[1]{#1}
\providecommand\showeprint[2][]{arXiv:#2}

\bibitem[Agrawal et~al\mbox{.}(2023)]%
        {agrawal2023enhancing}
\bibfield{author}{\bibinfo{person}{Sanjay Agrawal}, \bibinfo{person}{Srujana Merugu}, {and} \bibinfo{person}{Vivek Sembium}.} \bibinfo{year}{2023}\natexlab{}.
\newblock \showarticletitle{Enhancing e-commerce product search through reinforcement learning-powered query reformulation}. In \bibinfo{booktitle}{\emph{Proceedings of the 32nd ACM International Conference on Information and Knowledge Management}}. \bibinfo{pages}{4488--4494}.
\newblock


\bibitem[Antonellis et~al\mbox{.}(2008)]%
        {antonellis2008simrank}
\bibfield{author}{\bibinfo{person}{Ioannis Antonellis}, \bibinfo{person}{Hector Garcia-Molina}, {and} \bibinfo{person}{Chi-Chao Chang}.} \bibinfo{year}{2008}\natexlab{}.
\newblock \showarticletitle{Simrank++ query rewriting through link analysis of the clickgraph}. In \bibinfo{booktitle}{\emph{Proceedings of the 17th international conference on World Wide Web}}. \bibinfo{pages}{1177--1178}.
\newblock


\bibitem[Azar et~al\mbox{.}(2023)]%
        {IPO}
\bibfield{author}{\bibinfo{person}{Mohammad~Gheshlaghi Azar}, \bibinfo{person}{Mark Rowland}, \bibinfo{person}{Bilal Piot}, \bibinfo{person}{Daniel Guo}, \bibinfo{person}{Daniele Calandriello}, \bibinfo{person}{Michal Valko}, {and} \bibinfo{person}{Rémi Munos}.} \bibinfo{year}{2023}\natexlab{}.
\newblock \bibinfo{title}{A General Theoretical Paradigm to Understand Learning from Human Preferences}.
\newblock
\newblock
\showeprint[arxiv]{2310.12036}~[cs.AI]
\urldef\tempurl%
\url{https://arxiv.org/abs/2310.12036}
\showURL{%
\tempurl}


\bibitem[Bai et~al\mbox{.}(2023)]%
        {qwen}
\bibfield{author}{\bibinfo{person}{Jinze Bai}, \bibinfo{person}{Shuai Bai}, \bibinfo{person}{Yunfei Chu}, \bibinfo{person}{Zeyu Cui}, \bibinfo{person}{Kai Dang}, \bibinfo{person}{Xiaodong Deng}, \bibinfo{person}{Yang Fan}, \bibinfo{person}{Wenbin Ge}, \bibinfo{person}{Yu Han}, \bibinfo{person}{Fei Huang}, \bibinfo{person}{Binyuan Hui}, \bibinfo{person}{Luo Ji}, \bibinfo{person}{Mei Li}, \bibinfo{person}{Junyang Lin}, \bibinfo{person}{Runji Lin}, \bibinfo{person}{Dayiheng Liu}, \bibinfo{person}{Gao Liu}, \bibinfo{person}{Chengqiang Lu}, \bibinfo{person}{Keming Lu}, \bibinfo{person}{Jianxin Ma}, \bibinfo{person}{Rui Men}, \bibinfo{person}{Xingzhang Ren}, \bibinfo{person}{Xuancheng Ren}, \bibinfo{person}{Chuanqi Tan}, \bibinfo{person}{Sinan Tan}, \bibinfo{person}{Jianhong Tu}, \bibinfo{person}{Peng Wang}, \bibinfo{person}{Shijie Wang}, \bibinfo{person}{Wei Wang}, \bibinfo{person}{Shengguang Wu}, \bibinfo{person}{Benfeng Xu}, \bibinfo{person}{Jin Xu}, \bibinfo{person}{An Yang}, \bibinfo{person}{Hao Yang},
  \bibinfo{person}{Jian Yang}, \bibinfo{person}{Shusheng Yang}, \bibinfo{person}{Yang Yao}, \bibinfo{person}{Bowen Yu}, \bibinfo{person}{Hongyi Yuan}, \bibinfo{person}{Zheng Yuan}, \bibinfo{person}{Jianwei Zhang}, \bibinfo{person}{Xingxuan Zhang}, \bibinfo{person}{Yichang Zhang}, \bibinfo{person}{Zhenru Zhang}, \bibinfo{person}{Chang Zhou}, \bibinfo{person}{Jingren Zhou}, \bibinfo{person}{Xiaohuan Zhou}, {and} \bibinfo{person}{Tianhang Zhu}.} \bibinfo{year}{2023}\natexlab{}.
\newblock \bibinfo{title}{Qwen Technical Report}.
\newblock
\newblock
\showeprint[arxiv]{2309.16609}~[cs.CL]
\urldef\tempurl%
\url{https://arxiv.org/abs/2309.16609}
\showURL{%
\tempurl}


\bibitem[Casper et~al\mbox{.}(2023)]%
        {casper2023open}
\bibfield{author}{\bibinfo{person}{Stephen Casper}, \bibinfo{person}{Xander Davies}, \bibinfo{person}{Claudia Shi}, \bibinfo{person}{Thomas~Krendl Gilbert}, \bibinfo{person}{J{\'e}r{\'e}my Scheurer}, \bibinfo{person}{Javier Rando}, \bibinfo{person}{Rachel Freedman}, \bibinfo{person}{Tomasz Korbak}, \bibinfo{person}{David Lindner}, \bibinfo{person}{Pedro Freire}, {et~al\mbox{.}}} \bibinfo{year}{2023}\natexlab{}.
\newblock \showarticletitle{Open problems and fundamental limitations of reinforcement learning from human feedback}.
\newblock \bibinfo{journal}{\emph{arXiv preprint arXiv:2307.15217}} (\bibinfo{year}{2023}).
\newblock


\bibitem[Chen et~al\mbox{.}(2020)]%
        {chen2020rpm}
\bibfield{author}{\bibinfo{person}{Xiuying Chen}, \bibinfo{person}{Daorui Xiao}, \bibinfo{person}{Shen Gao}, \bibinfo{person}{Guojun Liu}, \bibinfo{person}{Wei Lin}, \bibinfo{person}{Bo Zheng}, \bibinfo{person}{Dongyan Zhao}, {and} \bibinfo{person}{Rui Yan}.} \bibinfo{year}{2020}\natexlab{}.
\newblock \showarticletitle{RPM-Oriented Query Rewriting Framework for E-commerce Keyword-Based Sponsored Search (Student Abstract)}. In \bibinfo{booktitle}{\emph{Proceedings of the AAAI Conference on Artificial Intelligence}}, Vol.~\bibinfo{volume}{34}. \bibinfo{pages}{13769--13770}.
\newblock


\bibitem[Cui et~al\mbox{.}(2002)]%
        {cui2002probabilistic}
\bibfield{author}{\bibinfo{person}{Hang Cui}, \bibinfo{person}{Ji-Rong Wen}, \bibinfo{person}{Jian-Yun Nie}, {and} \bibinfo{person}{Wei-Ying Ma}.} \bibinfo{year}{2002}\natexlab{}.
\newblock \showarticletitle{Probabilistic query expansion using query logs}. In \bibinfo{booktitle}{\emph{Proceedings of the 11th international conference on World Wide Web}}. \bibinfo{pages}{325--332}.
\newblock


\bibitem[Cui et~al\mbox{.}(2023)]%
        {chinese-llama-alpaca}
\bibfield{author}{\bibinfo{person}{Yiming Cui}, \bibinfo{person}{Ziqing Yang}, {and} \bibinfo{person}{Xin Yao}.} \bibinfo{year}{2023}\natexlab{}.
\newblock \showarticletitle{Efficient and Effective Text Encoding for Chinese LLaMA and Alpaca}.
\newblock \bibinfo{journal}{\emph{arXiv preprint arXiv:2304.08177}} (\bibinfo{year}{2023}).
\newblock
\urldef\tempurl%
\url{https://arxiv.org/abs/2304.08177}
\showURL{%
\tempurl}


\bibitem[Dai et~al\mbox{.}(2024)]%
        {dai2024enhancing}
\bibfield{author}{\bibinfo{person}{Aijun Dai}, \bibinfo{person}{Zhenyu Zhu}, \bibinfo{person}{Haiqing Hu}, \bibinfo{person}{Guoyu Tang}, \bibinfo{person}{Lin Liu}, {and} \bibinfo{person}{Sulong Xu}.} \bibinfo{year}{2024}\natexlab{}.
\newblock \showarticletitle{Enhancing E-Commerce Query Rewriting: A Large Language Model Approach with Domain-Specific Pre-Training and Reinforcement Learning}. In \bibinfo{booktitle}{\emph{Proceedings of the 33rd ACM International Conference on Information and Knowledge Management}}. \bibinfo{pages}{4439--4445}.
\newblock


\bibitem[Ethayarajh et~al\mbox{.}(2024)]%
        {KTO}
\bibfield{author}{\bibinfo{person}{Kawin Ethayarajh}, \bibinfo{person}{Winnie Xu}, \bibinfo{person}{Niklas Muennighoff}, \bibinfo{person}{Dan Jurafsky}, {and} \bibinfo{person}{Douwe Kiela}.} \bibinfo{year}{2024}\natexlab{}.
\newblock \bibinfo{title}{KTO: Model Alignment as Prospect Theoretic Optimization}.
\newblock
\newblock
\showeprint[arxiv]{2402.01306}~[cs.LG]
\urldef\tempurl%
\url{https://arxiv.org/abs/2402.01306}
\showURL{%
\tempurl}


\bibitem[Gao et~al\mbox{.}(2012)]%
        {gao2012learning}
\bibfield{author}{\bibinfo{person}{Jianfeng Gao}, \bibinfo{person}{Shasha Xie}, \bibinfo{person}{Xiaodong He}, {and} \bibinfo{person}{Alnur Ali}.} \bibinfo{year}{2012}\natexlab{}.
\newblock \showarticletitle{Learning lexicon models from search logs for query expansion}. In \bibinfo{booktitle}{\emph{Proceedings of EMNLP}}.
\newblock


\bibitem[Grbovic et~al\mbox{.}(2015)]%
        {grbovic2015context}
\bibfield{author}{\bibinfo{person}{Mihajlo Grbovic}, \bibinfo{person}{Nemanja Djuric}, \bibinfo{person}{Vladan Radosavljevic}, \bibinfo{person}{Fabrizio Silvestri}, {and} \bibinfo{person}{Narayan Bhamidipati}.} \bibinfo{year}{2015}\natexlab{}.
\newblock \showarticletitle{Context-and content-aware embeddings for query rewriting in sponsored search}. In \bibinfo{booktitle}{\emph{Proceedings of the 38th international ACM SIGIR conference on research and development in information retrieval}}. \bibinfo{pages}{383--392}.
\newblock


\bibitem[He et~al\mbox{.}(2016)]%
        {he2016learning}
\bibfield{author}{\bibinfo{person}{Yunlong He}, \bibinfo{person}{Jiliang Tang}, \bibinfo{person}{Hua Ouyang}, \bibinfo{person}{Changsung Kang}, \bibinfo{person}{Dawei Yin}, {and} \bibinfo{person}{Yi Chang}.} \bibinfo{year}{2016}\natexlab{}.
\newblock \showarticletitle{Learning to rewrite queries}. In \bibinfo{booktitle}{\emph{Proceedings of the 25th ACM International on Conference on Information and Knowledge Management}}. \bibinfo{pages}{1443--1452}.
\newblock


\bibitem[Jones et~al\mbox{.}(2006)]%
        {jones2006generating}
\bibfield{author}{\bibinfo{person}{Rosie Jones}, \bibinfo{person}{Benjamin Rey}, \bibinfo{person}{Omid Madani}, {and} \bibinfo{person}{Wiley Greiner}.} \bibinfo{year}{2006}\natexlab{}.
\newblock \showarticletitle{Generating query substitutions}. In \bibinfo{booktitle}{\emph{Proceedings of the 15th international conference on World Wide Web}}. \bibinfo{pages}{387--396}.
\newblock


\bibitem[Kaplan et~al\mbox{.}(2020)]%
        {kaplan2020scaling}
\bibfield{author}{\bibinfo{person}{Jared Kaplan}, \bibinfo{person}{Sam McCandlish}, \bibinfo{person}{Tom Henighan}, \bibinfo{person}{Tom~B Brown}, \bibinfo{person}{Benjamin Chess}, \bibinfo{person}{Rewon Child}, \bibinfo{person}{Scott Gray}, \bibinfo{person}{Alec Radford}, \bibinfo{person}{Jeffrey Wu}, {and} \bibinfo{person}{Dario Amodei}.} \bibinfo{year}{2020}\natexlab{}.
\newblock \showarticletitle{Scaling laws for neural language models}.
\newblock \bibinfo{journal}{\emph{arXiv preprint arXiv:2001.08361}} (\bibinfo{year}{2020}).
\newblock


\bibitem[Kenton and Toutanova(2019)]%
        {kenton2019bert}
\bibfield{author}{\bibinfo{person}{Jacob Devlin Ming-Wei~Chang Kenton} {and} \bibinfo{person}{Lee~Kristina Toutanova}.} \bibinfo{year}{2019}\natexlab{}.
\newblock \showarticletitle{Bert: Pre-training of deep bidirectional transformers for language understanding}. In \bibinfo{booktitle}{\emph{Proceedings of naacL-HLT}}, Vol.~\bibinfo{volume}{1}. Minneapolis, Minnesota, \bibinfo{pages}{2}.
\newblock


\bibitem[Lewis et~al\mbox{.}(2019)]%
        {bart}
\bibfield{author}{\bibinfo{person}{Mike Lewis}, \bibinfo{person}{Yinhan Liu}, \bibinfo{person}{Naman Goyal}, \bibinfo{person}{Marjan Ghazvininejad}, \bibinfo{person}{Abdelrahman Mohamed}, \bibinfo{person}{Omer Levy}, \bibinfo{person}{Ves Stoyanov}, {and} \bibinfo{person}{Luke Zettlemoyer}.} \bibinfo{year}{2019}\natexlab{}.
\newblock \bibinfo{title}{BART: Denoising Sequence-to-Sequence Pre-training for Natural Language Generation, Translation, and Comprehension}.
\newblock
\newblock
\showeprint[arxiv]{1910.13461}~[cs.CL]
\urldef\tempurl%
\url{https://arxiv.org/abs/1910.13461}
\showURL{%
\tempurl}


\bibitem[Li et~al\mbox{.}(2021)]%
        {MORLHF}
\bibfield{author}{\bibinfo{person}{Kaiwen Li}, \bibinfo{person}{Tao Zhang}, {and} \bibinfo{person}{Rui Wang}.} \bibinfo{year}{2021}\natexlab{}.
\newblock \showarticletitle{Deep Reinforcement Learning for Multiobjective Optimization}.
\newblock \bibinfo{journal}{\emph{IEEE Transactions on Cybernetics}} \bibinfo{volume}{51}, \bibinfo{number}{6} (\bibinfo{date}{June} \bibinfo{year}{2021}), \bibinfo{pages}{3103–3114}.
\newblock
\showISSN{2168-2275}
\urldef\tempurl%
\url{https://doi.org/10.1109/tcyb.2020.2977661}
\showDOI{\tempurl}


\bibitem[Li et~al\mbox{.}(2022)]%
        {li2022query}
\bibfield{author}{\bibinfo{person}{Sen Li}, \bibinfo{person}{Fuyu Lv}, \bibinfo{person}{Taiwei Jin}, \bibinfo{person}{Guiyang Li}, \bibinfo{person}{Yukun Zheng}, \bibinfo{person}{Tao Zhuang}, \bibinfo{person}{Qingwen Liu}, \bibinfo{person}{Xiaoyi Zeng}, \bibinfo{person}{James Kwok}, {and} \bibinfo{person}{Qianli Ma}.} \bibinfo{year}{2022}\natexlab{}.
\newblock \showarticletitle{Query Rewriting in TaoBao Search}. In \bibinfo{booktitle}{\emph{Proceedings of the 31st ACM International Conference on Information \& Knowledge Management}}. \bibinfo{pages}{3262--3271}.
\newblock


\bibitem[Lian et~al\mbox{.}(2019)]%
        {lian2019end}
\bibfield{author}{\bibinfo{person}{Yijiang Lian}, \bibinfo{person}{Zhijie Chen}, \bibinfo{person}{Jinlong Hu}, \bibinfo{person}{Kefeng Zhang}, \bibinfo{person}{Chunwei Yan}, \bibinfo{person}{Muchenxuan Tong}, \bibinfo{person}{Wenying Han}, \bibinfo{person}{Hanju Guan}, \bibinfo{person}{Ying Li}, \bibinfo{person}{Ying Cao}, {et~al\mbox{.}}} \bibinfo{year}{2019}\natexlab{}.
\newblock \showarticletitle{An end-to-end Generative Retrieval Method for Sponsored Search Engine--Decoding Efficiently into a Closed Target Domain}.
\newblock \bibinfo{journal}{\emph{arXiv preprint arXiv:1902.00592}} (\bibinfo{year}{2019}).
\newblock


\bibitem[Liu et~al\mbox{.}(1907)]%
        {liu1907roberta}
\bibfield{author}{\bibinfo{person}{Y Liu}, \bibinfo{person}{M Ott}, \bibinfo{person}{N Goyal}, \bibinfo{person}{J Du}, \bibinfo{person}{M Joshi}, \bibinfo{person}{D Chen}, \bibinfo{person}{O Levy}, \bibinfo{person}{M Lewis}, \bibinfo{person}{L Zettlemoyer}, {and} \bibinfo{person}{V Stoyanov}.} \bibinfo{year}{1907}\natexlab{}.
\newblock \showarticletitle{RoBERTa: A robustly optimized BERT pretraining approach. arXiv [Preprint](2019)}.
\newblock \bibinfo{journal}{\emph{arXiv preprint arXiv:1907.11692}} (\bibinfo{year}{1907}).
\newblock


\bibitem[Loshchilov and Hutter(2019)]%
        {adamw2019decoupledweightdecayregularization}
\bibfield{author}{\bibinfo{person}{Ilya Loshchilov} {and} \bibinfo{person}{Frank Hutter}.} \bibinfo{year}{2019}\natexlab{}.
\newblock \bibinfo{title}{Decoupled Weight Decay Regularization}.
\newblock
\newblock
\showeprint[arxiv]{1711.05101}~[cs.LG]
\urldef\tempurl%
\url{https://arxiv.org/abs/1711.05101}
\showURL{%
\tempurl}


\bibitem[Ma et~al\mbox{.}(2023)]%
        {ma2023query}
\bibfield{author}{\bibinfo{person}{Xinbei Ma}, \bibinfo{person}{Yeyun Gong}, \bibinfo{person}{Pengcheng He}, \bibinfo{person}{Hai Zhao}, {and} \bibinfo{person}{Nan Duan}.} \bibinfo{year}{2023}\natexlab{}.
\newblock \showarticletitle{Query rewriting for retrieval-augmented large language models}.
\newblock \bibinfo{journal}{\emph{arXiv preprint arXiv:2305.14283}} (\bibinfo{year}{2023}).
\newblock


\bibitem[Meng et~al\mbox{.}(2024)]%
        {SimPO}
\bibfield{author}{\bibinfo{person}{Yu Meng}, \bibinfo{person}{Mengzhou Xia}, {and} \bibinfo{person}{Danqi Chen}.} \bibinfo{year}{2024}\natexlab{}.
\newblock \bibinfo{title}{SimPO: Simple Preference Optimization with a Reference-Free Reward}.
\newblock
\newblock
\showeprint[arxiv]{2405.14734}~[cs.CL]
\urldef\tempurl%
\url{https://arxiv.org/abs/2405.14734}
\showURL{%
\tempurl}


\bibitem[Mohankumar et~al\mbox{.}(2021)]%
        {mohankumar2021diversity}
\bibfield{author}{\bibinfo{person}{Akash~Kumar Mohankumar}, \bibinfo{person}{Nikit Begwani}, {and} \bibinfo{person}{Amit Singh}.} \bibinfo{year}{2021}\natexlab{}.
\newblock \showarticletitle{Diversity driven query rewriting in search advertising}. In \bibinfo{booktitle}{\emph{Proceedings of the 27th ACM SIGKDD Conference on Knowledge Discovery \& Data Mining}}. \bibinfo{pages}{3423--3431}.
\newblock


\bibitem[Ouyang et~al\mbox{.}(2024)]%
        {instructGPT}
\bibfield{author}{\bibinfo{person}{Long Ouyang}, \bibinfo{person}{Jeff Wu}, \bibinfo{person}{Xu Jiang}, \bibinfo{person}{Diogo Almeida}, \bibinfo{person}{Carroll~L. Wainwright}, \bibinfo{person}{Pamela Mishkin}, \bibinfo{person}{Chong Zhang}, \bibinfo{person}{Sandhini Agarwal}, \bibinfo{person}{Katarina Slama}, \bibinfo{person}{Alex Ray}, \bibinfo{person}{John Schulman}, \bibinfo{person}{Jacob Hilton}, \bibinfo{person}{Fraser Kelton}, \bibinfo{person}{Luke Miller}, \bibinfo{person}{Maddie Simens}, \bibinfo{person}{Amanda Askell}, \bibinfo{person}{Peter Welinder}, \bibinfo{person}{Paul Christiano}, \bibinfo{person}{Jan Leike}, {and} \bibinfo{person}{Ryan Lowe}.} \bibinfo{year}{2024}\natexlab{}.
\newblock \showarticletitle{Training language models to follow instructions with human feedback}. In \bibinfo{booktitle}{\emph{Proceedings of the 36th International Conference on Neural Information Processing Systems}} (New Orleans, LA, USA) \emph{(\bibinfo{series}{NIPS '22})}. \bibinfo{publisher}{Curran Associates Inc.}, \bibinfo{address}{Red Hook, NY, USA}, Article \bibinfo{articleno}{2011}, \bibinfo{numpages}{15}~pages.
\newblock
\showISBNx{9781713871088}


\bibitem[Pang et~al\mbox{.}(2025)]%
        {pang2025generative}
\bibfield{author}{\bibinfo{person}{Ming Pang}, \bibinfo{person}{Chunyuan Yuan}, \bibinfo{person}{Xiaoyu He}, \bibinfo{person}{Zheng Fang}, \bibinfo{person}{Donghao Xie}, \bibinfo{person}{Fanyi Qu}, \bibinfo{person}{Xue Jiang}, \bibinfo{person}{Changping Peng}, \bibinfo{person}{Zhangang Lin}, \bibinfo{person}{Zheng Luo}, {et~al\mbox{.}}} \bibinfo{year}{2025}\natexlab{}.
\newblock \showarticletitle{Generative Retrieval and Alignment Model: A New Paradigm for E-commerce Retrieval}.
\newblock \bibinfo{journal}{\emph{arXiv preprint arXiv:2504.01403}} (\bibinfo{year}{2025}).
\newblock


\bibitem[Peng et~al\mbox{.}(2024)]%
        {peng2024large}
\bibfield{author}{\bibinfo{person}{Wenjun Peng}, \bibinfo{person}{Guiyang Li}, \bibinfo{person}{Yue Jiang}, \bibinfo{person}{Zilong Wang}, \bibinfo{person}{Dan Ou}, \bibinfo{person}{Xiaoyi Zeng}, \bibinfo{person}{Derong Xu}, \bibinfo{person}{Tong Xu}, {and} \bibinfo{person}{Enhong Chen}.} \bibinfo{year}{2024}\natexlab{}.
\newblock \showarticletitle{Large language model based long-tail query rewriting in taobao search}. In \bibinfo{booktitle}{\emph{Companion Proceedings of the ACM on Web Conference 2024}}. \bibinfo{pages}{20--28}.
\newblock


\bibitem[Rafailov et~al\mbox{.}(2024)]%
        {rafailov2024direct}
\bibfield{author}{\bibinfo{person}{Rafael Rafailov}, \bibinfo{person}{Archit Sharma}, \bibinfo{person}{Eric Mitchell}, \bibinfo{person}{Christopher~D Manning}, \bibinfo{person}{Stefano Ermon}, {and} \bibinfo{person}{Chelsea Finn}.} \bibinfo{year}{2024}\natexlab{}.
\newblock \showarticletitle{Direct preference optimization: Your language model is secretly a reward model}.
\newblock \bibinfo{journal}{\emph{Advances in Neural Information Processing Systems}}  \bibinfo{volume}{36} (\bibinfo{year}{2024}).
\newblock


\bibitem[Riezler and Liu(2010)]%
        {riezler2010query}
\bibfield{author}{\bibinfo{person}{Stefan Riezler} {and} \bibinfo{person}{Yi Liu}.} \bibinfo{year}{2010}\natexlab{}.
\newblock \showarticletitle{Query rewriting using monolingual statistical machine translation}.
\newblock \bibinfo{journal}{\emph{Computational Linguistics}} \bibinfo{volume}{36}, \bibinfo{number}{3} (\bibinfo{year}{2010}), \bibinfo{pages}{569--582}.
\newblock


\bibitem[Schulman et~al\mbox{.}(2017)]%
        {PPO}
\bibfield{author}{\bibinfo{person}{John Schulman}, \bibinfo{person}{Filip Wolski}, \bibinfo{person}{Prafulla Dhariwal}, \bibinfo{person}{Alec Radford}, {and} \bibinfo{person}{Oleg Klimov}.} \bibinfo{year}{2017}\natexlab{}.
\newblock \bibinfo{title}{Proximal Policy Optimization Algorithms}.
\newblock
\newblock
\showeprint[arxiv]{1707.06347}~[cs.LG]
\urldef\tempurl%
\url{https://arxiv.org/abs/1707.06347}
\showURL{%
\tempurl}


\bibitem[Song et~al\mbox{.}(2021)]%
        {song2021triangular}
\bibfield{author}{\bibinfo{person}{Zhenqiao Song}, \bibinfo{person}{Jiaze Chen}, \bibinfo{person}{Hao Zhou}, {and} \bibinfo{person}{Lei Li}.} \bibinfo{year}{2021}\natexlab{}.
\newblock \showarticletitle{Triangular Bidword Generation for Sponsored Search Auction}. In \bibinfo{booktitle}{\emph{Proceedings of the 14th ACM International Conference on Web Search and Data Mining}}. \bibinfo{pages}{707--715}.
\newblock


\bibitem[Sordoni et~al\mbox{.}(2015)]%
        {sordoni2015hierarchical}
\bibfield{author}{\bibinfo{person}{Alessandro Sordoni}, \bibinfo{person}{Yoshua Bengio}, \bibinfo{person}{Hossein Vahabi}, \bibinfo{person}{Christina Lioma}, \bibinfo{person}{Jakob Grue~Simonsen}, {and} \bibinfo{person}{Jian-Yun Nie}.} \bibinfo{year}{2015}\natexlab{}.
\newblock \showarticletitle{A hierarchical recurrent encoder-decoder for generative context-aware query suggestion}. In \bibinfo{booktitle}{\emph{proceedings of the 24th ACM international on conference on information and knowledge management}}. \bibinfo{pages}{553--562}.
\newblock


\bibitem[Touvron et~al\mbox{.}(2023)]%
        {llama2}
\bibfield{author}{\bibinfo{person}{Hugo Touvron}, \bibinfo{person}{Louis Martin}, \bibinfo{person}{Kevin Stone}, \bibinfo{person}{Peter Albert}, {and} \bibinfo{person}{Amjad Almahairi}.} \bibinfo{year}{2023}\natexlab{}.
\newblock \bibinfo{title}{Llama 2: Open Foundation and Fine-Tuned Chat Models}.
\newblock
\newblock
\showeprint[arxiv]{2307.09288}~[cs.CL]
\urldef\tempurl%
\url{https://arxiv.org/abs/2307.09288}
\showURL{%
\tempurl}


\bibitem[Vakulenko et~al\mbox{.}(2021)]%
        {vakulenko2021question}
\bibfield{author}{\bibinfo{person}{Svitlana Vakulenko}, \bibinfo{person}{Shayne Longpre}, \bibinfo{person}{Zhucheng Tu}, {and} \bibinfo{person}{Raviteja Anantha}.} \bibinfo{year}{2021}\natexlab{}.
\newblock \showarticletitle{Question rewriting for conversational question answering}. In \bibinfo{booktitle}{\emph{Proceedings of the 14th ACM international conference on web search and data mining}}. \bibinfo{pages}{355--363}.
\newblock


\bibitem[Vaswani(2017)]%
        {vaswani2017attention}
\bibfield{author}{\bibinfo{person}{A Vaswani}.} \bibinfo{year}{2017}\natexlab{}.
\newblock \showarticletitle{Attention is all you need}.
\newblock \bibinfo{journal}{\emph{Advances in Neural Information Processing Systems}} (\bibinfo{year}{2017}).
\newblock


\bibitem[Wang et~al\mbox{.}(2021)]%
        {wang2021queen}
\bibfield{author}{\bibinfo{person}{Yaxuan Wang}, \bibinfo{person}{Hanqing Lu}, \bibinfo{person}{Yunwen Xu}, \bibinfo{person}{Rahul Goutam}, \bibinfo{person}{Yiwei Song}, {and} \bibinfo{person}{Bing Yin}.} \bibinfo{year}{2021}\natexlab{}.
\newblock \showarticletitle{QUEEN: Neural query rewriting in e-commerce}. In \bibinfo{booktitle}{\emph{The Web Conference 2021}}.
\newblock
\urldef\tempurl%
\url{https://www.amazon.science/publications/queen-neural-query-rewriting-in-e-commerce}
\showURL{%
\tempurl}


\bibitem[Wei et~al\mbox{.}(2022)]%
        {wei2022emergent}
\bibfield{author}{\bibinfo{person}{Jason Wei}, \bibinfo{person}{Yi Tay}, \bibinfo{person}{Rishi Bommasani}, \bibinfo{person}{Colin Raffel}, \bibinfo{person}{Barret Zoph}, \bibinfo{person}{Sebastian Borgeaud}, \bibinfo{person}{Dani Yogatama}, \bibinfo{person}{Maarten Bosma}, \bibinfo{person}{Denny Zhou}, \bibinfo{person}{Donald Metzler}, {et~al\mbox{.}}} \bibinfo{year}{2022}\natexlab{}.
\newblock \showarticletitle{Emergent abilities of large language models}.
\newblock \bibinfo{journal}{\emph{arXiv preprint arXiv:2206.07682}} (\bibinfo{year}{2022}).
\newblock


\bibitem[Yuan et~al\mbox{.}(2024)]%
        {yuan2024semi}
\bibfield{author}{\bibinfo{person}{Chunyuan Yuan}, \bibinfo{person}{Ming Pang}, \bibinfo{person}{Zheng Fang}, \bibinfo{person}{Xue Jiang}, \bibinfo{person}{Changping Peng}, {and} \bibinfo{person}{Zhangang Lin}.} \bibinfo{year}{2024}\natexlab{}.
\newblock \showarticletitle{A Semi-supervised Multi-channel Graph Convolutional Network for Query Classification in E-commerce}. In \bibinfo{booktitle}{\emph{Companion Proceedings of the ACM Web Conference 2024}}. \bibinfo{pages}{56--64}.
\newblock


\bibitem[Yuan et~al\mbox{.}(2023)]%
        {yuan2023multi}
\bibfield{author}{\bibinfo{person}{Chunyuan Yuan}, \bibinfo{person}{Yiming Qiu}, \bibinfo{person}{Mingming Li}, \bibinfo{person}{Haiqing Hu}, \bibinfo{person}{Songlin Wang}, {and} \bibinfo{person}{Sulong Xu}.} \bibinfo{year}{2023}\natexlab{}.
\newblock \showarticletitle{A multi-granularity matching attention network for query intent classification in e-commerce retrieval}. In \bibinfo{booktitle}{\emph{Companion Proceedings of the ACM Web Conference 2023}}. \bibinfo{pages}{416--420}.
\newblock


\bibitem[Zhou et~al\mbox{.}(2024)]%
        {MODPO}
\bibfield{author}{\bibinfo{person}{Zhanhui Zhou}, \bibinfo{person}{Jie Liu}, \bibinfo{person}{Jing Shao}, \bibinfo{person}{Xiangyu Yue}, \bibinfo{person}{Chao Yang}, \bibinfo{person}{Wanli Ouyang}, {and} \bibinfo{person}{Yu Qiao}.} \bibinfo{year}{2024}\natexlab{}.
\newblock \bibinfo{title}{Beyond One-Preference-Fits-All Alignment: Multi-Objective Direct Preference Optimization}.
\newblock
\newblock
\showeprint[arxiv]{2310.03708}~[cs.LG]
\urldef\tempurl%
\url{https://arxiv.org/abs/2310.03708}
\showURL{%
\tempurl}


\bibitem[Zuo et~al\mbox{.}(2023)]%
        {zuo2023industry}
\bibfield{author}{\bibinfo{person}{Simiao Zuo}, \bibinfo{person}{Qingyu Yin}, \bibinfo{person}{Haoming Jiang}, \bibinfo{person}{Shaohui Xi}, \bibinfo{person}{Bing Yin}, \bibinfo{person}{Chao Zhang}, {and} \bibinfo{person}{Tuo Zhao}.} \bibinfo{year}{2023}\natexlab{}.
\newblock \showarticletitle{Context-Aware Query Rewriting for Improving Users' Search Experience on E-commerce Websites}. In \bibinfo{booktitle}{\emph{The 61st Annual Meeting Of The Association For Computational Linguistics}}.
\newblock


\end{thebibliography}

\end{document}